\documentclass{article}
\pdfoutput=1
\usepackage{PRIMEarxiv}
\usepackage{amsmath,amssymb,amsfonts}
\usepackage{algorithmic}
\usepackage{algorithm}
\usepackage{graphicx}
\usepackage{booktabs}
\usepackage{tablefootnote}
\usepackage{threeparttable}
\usepackage[switch, pagewise]{lineno}  
\usepackage[authoryear]{natbib}
\usepackage{stfloats}  
\usepackage{hyperref}

\begin{document}
\title{LMDA-Net:A lightweight multi-dimensional attention network for general EEG-based brain-computer interface paradigms and interpretability}


\author{
    Zhengqing Miao \footnotemark[1] \And
    Xin Zhang      \footnotemark[2] \And
    Meirong Zhao   \footnotemark[1] \And
    Dong Ming      \footnotemark[2]
    }
\renewcommand{\thefootnote}{\fnsymbol{footnote}}
\footnotetext{Email: Zhengqing Miao(mzq@tju.edu.cn)}
\footnotetext{Corresponding author Meirong Zhao (meirongzhao@tju.edu.cn) and Xin Zhang (xin\_zhang\_bme@tju.edu.cn)}
\footnotetext[1]{State Key Laboratory of Precision Measuring Technology and Instruments, School of Precision Instrument and Opto-electronics Engineering, Tianjin University, Tianjin 300072, China.}
\footnotetext[2]{Laboratory of Neural Engineering and Rehabilitation, Department of Biomedical Engineering, School of Precision Instruments and Optoelectronics Engineering, Tianjin University, and also with the Tianjin International Joint Research Center for Neural Engineering, Academy of Medical Engineering and Translational Medicine, Tianjin University, Tianjin 300072, China}
            


\maketitle

\begin{abstract}
Electroencephalography (EEG)-based brain-computer interfaces (BCIs) pose a challenge for decoding due to their low spatial resolution and signal-to-noise ratio. Typically, EEG-based recognition of activities and states involves the use of prior neuroscience knowledge to generate quantitative EEG features, which may limit BCI performance. Although neural network-based methods can effectively extract features, they often encounter issues such as poor generalization across datasets, high predicting volatility, and low model interpretability.
To address these limitations, we propose a novel lightweight multi-dimensional attention network, called LMDA-Net. By incorporating two novel attention modules designed specifically for EEG signals, the channel attention module and the depth attention module, LMDA-Net is able to effectively integrate features from multiple dimensions, resulting in improved classification performance across various BCI tasks.
LMDA-Net was evaluated on four high-impact public datasets, including motor imagery (MI) and P300-Speller paradigms, and was compared with other representative models. The experimental results demonstrate that LMDA-Net outperforms other representative methods in terms of classification accuracy and predicting volatility, achieving the highest accuracy in all datasets within 300 training epochs. Ablation experiments further confirm the effectiveness of the channel attention module and the depth attention module.
To facilitate an in-depth understanding of the features extracted by LMDA-Net, we propose class-specific neural network feature interpretability algorithms that are suitable for event-related potentials (ERPs) and event-related desynchronization/synchronization (ERD/ERS). By mapping the output of the specific layer of MDA-Net to the time or spatial domain through class activation maps, the resulting feature visualizations can provide interpretable analysis and establish connections with EEG time-spatial analysis in neuroscience. In summary, LMDA-Net shows great potential as a general online decoding model for various EEG tasks. 
\end{abstract}


\section{Introduction}
\label{sec:introduction}
Electroencephalography (EEG)-based brain-computer interfaces (BCIs) have emerged as powerful tools for various applications, such as motor rehabilitation [\cite{lazarou2018eeg}], brain function regulation [\cite{sterman1996physiological}], psychiatric disorders monitoring [\cite{pollock1990quantitative}], and entertainment [\cite{olfers2018game}]. Non-invasive EEG has gained widespread use due to its ease of signal acquisition and high temporal resolution. However, the low signal-to-noise ratio and poor spatial resolution of EEG present significant challenges for the development of effective EEG-based BCIs. Furthermore, EEG is often acquired with different paradigms [\cite{lakshmi2014survey}] that utilize distinct neuroscientific principles, resulting in various signal features across different signal domains.
For instance, motor imagery (MI) tasks can evoke event-related synchronization/desynchronization (ERD/ERS) in the Mu frequency bands, which exhibit strong time-frequency characteristics [\cite{pfurtscheller2006mu}]. The filter bank common spatial pattern (FBCSP) method [\cite{ang2008filter}] is a common technique for extracting features from MI data. On the other hand, event-related potentials (ERPs), such as the P300, can induce time-locked and phase-locked changes in the activity of neuronal populations [\cite{pfurtscheller2006mu, david2006mechanisms}]. To enhance the signal-to-noise ratio of ERPs, averaging techniques or spatial filters such as Xdawn [\cite{rivet2009xdawn}] are often used.

\begin{figure*}
\centerline{\includegraphics[width=\textwidth]{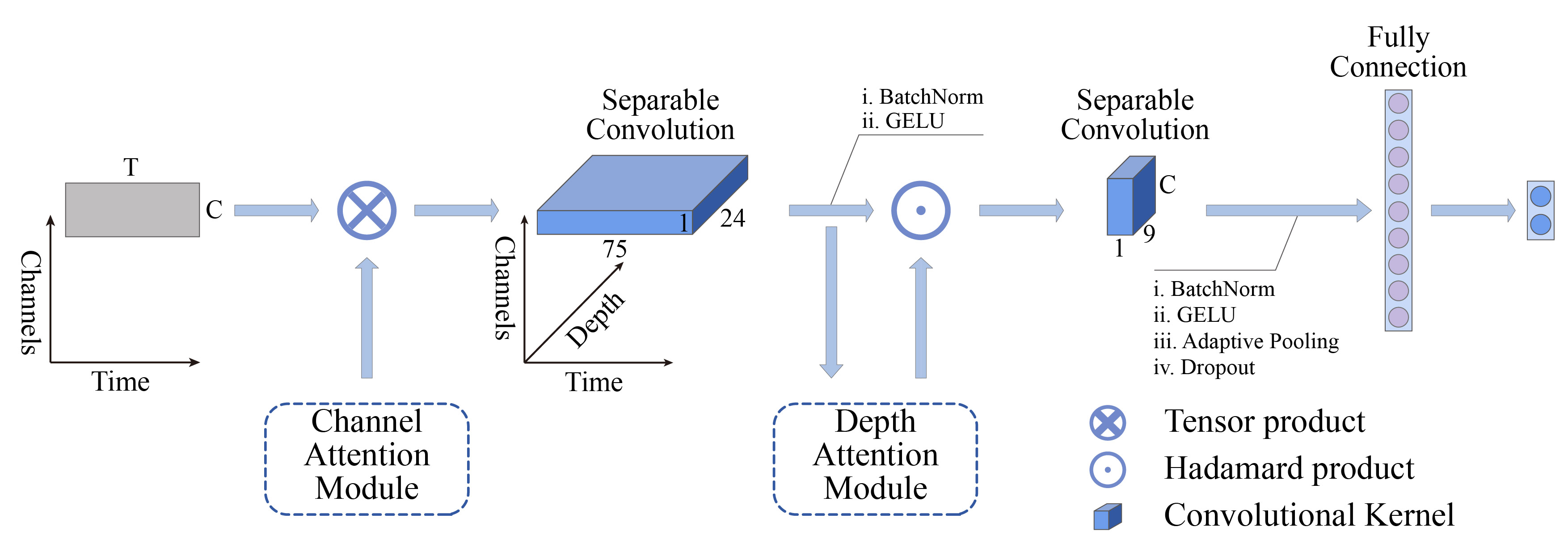}}
\caption{The architecture of LMDA-Net, where C is the number of EEG channels, T is the number of time samples in a trial. Other parameters are also shown in the figure. The output of LMDA-Net depends on the number of task categories. LMDA-Net includes three parts: the benchmark network (see \ref{benchmark}), channel attention module (see \ref{channel_attention}) and depth attention module (see \ref{depth_attention}). Two attention modules are used to assist the benchmark network in feature extraction, where channel attention module is introduced by Tensor product and depth attention module is introduced by Hadamard product of tensors.}
\label{fig:lmda}
\end{figure*}

In brain-computer interface (BCI) research, developing task- or paradigm-specific feature algorithms is labor-intensive. To address this challenge, researchers have explored the use of neural network technologies [\cite{schwemmer2018meeting, craik2019deep, al2021deep,schirrmeister2017deep, lawhern2018eegnet}]. For instance, \cite{schirrmeister2017deep} demonstrated that a shallow network with two EEG-specific convolutional layers can achieve comparable classification performance to filter bank common spatial pattern (FBCSP) in motor imagery (MI) signals. EEGNet [\cite{lawhern2018eegnet}] achieved good performance on EEG data from different paradigms, especially for event-related potentials (ERPs), by designing a lightweight network with two temporal convolutional layers. However, the special characteristics of EEG signals pose challenges for deep neural network architectures that are typically designed with convolutional layers in the time and spatial dimensions. This dimension-specific convolutional design may limit the ability of neural networks to extract high-dimensional features. Additionally, EEG data are often scarce for individual participants, and sharing data among participants is difficult due to ethical and privacy concerns, which further constrains deep neural network design. 

Attention mechanism is a research hotspot in the field of deep learning, which can facilitate high-value information screening from a large amount of information [\cite{mnih2014recurrent}]. 
There are generally three types of attention mechanism used in the literature: global attention mechanism [\cite{bahdanau2014neural, xu2015show}], local attention mechanism [\cite{luong2015effective}], and self-attention mechanism [\cite{dosovitskiy2020image, vaswani2017attention}]. In the field of EEG decoding, attention mechanisms have also been applied in recent studies. For example, \cite{xie2022transformer} proposed a transformer-based model combined with a convolutional neural network for MI-EEG classification, while \cite{zheng2022copula} proposed a copula-based transformer for assessing visual discomfort induced by stereoscopic 3D. Additionally, \cite{bagchi2022eeg} proposed a network incorporating multi-headed self-attention and temporal convolution for single-trial EEG-based visual stimulus classification.
However, the application of existing attention mechanisms in EEG signals faces three key challenges due to the specific characteristics of EEG.
Firstly, the above-mentioned attention mechanisms require a huge number of parameters to be learned, especially the self-attention mechanism based methods, like transformers [\cite{vaswani2017attention}]. This makes it difficult to effectively train millions or billions of parameters with only a few hundred EEG samples. 
Secondly, although raw EEG is a two-dimensional (time and spatial) signal, the information contained in the time and spatial domains is quite different and should not be treated equally. The importance of EEG channels varies significantly based on prior knowledge from neuroscience [\cite{arvaneh2011optimizing}], which makes image-based attention modules difficult to transfer to EEG decoding due to the dimension inequality (the length and width of the image are of equal weights). 
Thirdly, EEG signals have a lower signal-to-noise ratio compared to images, making it essential to consider the influence of noise when applying attention mechanisms in EEG decoding.

To solve these issues, in this paper, we propose a unified Lightweight neural network model integrated with Multi-Dimensional Attention modules (LMDA-Net in the following context), for EEG recognition from different paradigms, different acquisition systems.
The LMDA-Net contains 3 parts:

\begin{enumerate}
\item	Benchmark network: a lightweight shallow network, mainly responsible for EEG feature extraction and classification.
\item	Channel attention module: a novel attention module for improving the spatial resolution of raw EEG signals suitable for feature extraction in neural networks.
\item	Depth attention module: a depth information screening module to strengthen the transformation of high-dimensional EEG features in various dimensions.
\end{enumerate}

Neural network models are often regarded as black boxes, making it difficult for humans to understand what they have learned from labeled EEG data. To address this issue, several studies have used feature visualization techniques to represent the feature distributions and spatial topologies of different classes or network units.
\cite{zhao2020deep} and \cite{miao2022priming} visualized the feature distribution of different classes by t-SNE [\cite{van2008visualizing}]. 
\cite{lawhern2018eegnet} visualized the features derived from the convolution kernel and showed the spatial topoplots. 
\cite{schirrmeister2017deep} visualized the correlation difference between the input-feature and unit-output of the network.
\cite{tsception2022} visualized the area of interest of the model and ploted the topological map. 
 However, there is still a gap between the feature visualization and interpretability of these methods.
While feature visualization has been widely used in computer vision [\cite{guidotti2018survey}, \cite{selvaraju2017grad}], applying this technique to EEG signals presents unique challenges. Compared to images, EEG signals have a low signal-to-noise ratio, which means that the features learned by the network model may contain noise. Moreover, visualized features for EEG cannot be interpreted directly as they can for images, making it difficult to compare the results with prior knowledge.
 
To address these challenges, we propose novel feature visualization algorithms for ERPs and ERD/ERS, respectively. We first use Eigen-CAM [\cite{muhammad2020eigen}] to visualize the principle components of the learned features from LMDA-Net. Next, we map the resulting visualizations to the time and spatial domains, thus providing an interpretation of the feature visualization results by leveraging the findings of the EEG time-spatial analysis as prior knowledge.

The major contributions of this paper can be summarized as follows. 

\begin{enumerate}
\item	A lightweight network with multi-dimensional attention module for various EEG tasks analysis is proposed and validated, where a channel attention module is introduced, specifically for EEG, to screen the spatial information and a depth attention module is introduced to enhance the feature integration of the time domain and the spatial domain. 
\item	The classification performance of LMDA-Net is evaluated on four high-impact public EEG datasets, including three MI datasets of different acquisition setups and one P300-Speller dataset. The experimental results show that in all four datasets, LMDA-Net achieves the highest classification accuracy campared with other representative methods.  Also LMDA-Net requires no more than 300 epochs of training, which is probably the easiest model to train among existing end-to-end neural network models of EEG decoding. Furthermore, the predicting volatility of LMDA-Net is also very low, which shows great potential for practical online deployment.
\item	Class-specific neural network feature interpretability algorithms are proposed for ERPs and ERD/ERS, respectively. By remapping class-specific features learned by LMDA-Net to the time and spatial domains, conclusions of EEG time-spatial analysis in the literatures for both P300-Speller and MI prove the rationality of the features learned by LMDA-Net.
\end{enumerate}

\section{Material and methods}
\begin{table*}
\caption{Summary of the public datasets used in the experiments}
\label{table:datasets}
\begin{threeparttable}
\begin{tabular}{llllllllllll}
\toprule
\textbf{Datasets} & \textbf{\begin{tabular}[c]{@{}l@{}}\# of \\ channels\end{tabular}} & \textbf{\begin{tabular}[c]{@{}l@{}}\# of \\ participants\end{tabular}} & \textbf{\begin{tabular}[c]{@{}l@{}}\# of \\ trials\end{tabular}} & \textbf{\begin{tabular}[c]{@{}l@{}}\# of \\ classes\end{tabular}} & \textbf{Test metrics} & \textbf{\begin{tabular}[c]{@{}l@{}}Sampling\\ rate(HZ)\end{tabular}} & \textbf{\begin{tabular}[c]{@{}l@{}}Duration\\ (s)\end{tabular}} & \textbf{\begin{tabular}[c]{@{}l@{}}Data\\  split\end{tabular}} \\ 
\midrule
BCI4-2A           & 22                                                                 & 9                                                                  & 576                                                                        & 4                                                                 & ACC, kappa            & 250                                                              & 4                                                                     & Official\tnote{1}                                                                \\
BCI4-2B           & 3                                                                  & 9                                                                  & 720                                                                        & 2                                                                 & ACC, kappa            & 250                                                              & 4                                                                     & Official                                                                \\
BCI3-4a           & 118                                                                & 5                                                                  & 280                                                                        & 2                                                                 & ACC, kappa            & 1000                                                             & 3.5                                                                   & Cross\tnote{2}                                                                    \\
Kaggle-ERN        & 56                                                                 & 26                                                                 & 340                                                                        & 2                                                                 & AUC                   & 600                                                              & 1.25                                                                  & Official                                                       \\ 
\bottomrule
\end{tabular}
 \begin{tablenotes}
        \footnotesize
        \item[1] Official indicates that the division of the training set and the test set is based on the annotation of the dataset itself.
        \item[2] Cross indicates using data from other participants to train a participant-agnostic model.
      \end{tablenotes}
    \end{threeparttable}
\end{table*}

\subsection{Data}
To verify the performance of LMDA-Net, four high-impact public datasets, which were recorded under different acquisition paradigms, different numbers of tasks, and different numbers of channels, were selected. These datasets also contained within-participant and cross-participant conditions. 
The basic information of the dataset is shown in Table \ref{table:datasets}. 

ERPs and ERD/ERS are two common responses of neuronal structures recorded by EEG [\cite{david2006mechanisms}.]
One of the most studied  paradigms for ERD/ERS is motor imagery (MI), which either decreases or increases of power in specific frequency bands during the activity. Signals of MI are often characterized in the time-frequency domain and have a strong oscillating property [\cite{pfurtscheller1999event}]. In view of the complexity of decoding this kind of signal, we selected 3 public datasets of motor imagery with significant differences for experiments. ERPs related activity is often characterized in the time domain. This kind of activity is time-locked and phased-locked, and is generally robust across participants and contain well-stereotyped waveforms [\cite{lawhern2018eegnet}, \cite{polich2007updating}]. To this end, we selected a public dataset of P300-speller with great influence for research.

\subsubsection{Dataset 1: MI (BCI4-2A)}
In BCI4-2A dataset\footnote{\url{www.bbci.de/competition/iv/\#dataset2a}}, EEG was collected from a 10-20 system with 22 EEG channels at a sampling rate of 250 Hz from nine healthy participants (ID A01-A09) in two different sessions. Each participant participated in four motor imagery tasks, including imagining the movement of left hand, right hand, both feet and tongue. Each session contains 288 trials of EEG data. All data collected in the first session were used for training and those in the second session were used for test. According to \cite{schirrmeister2017deep}, temporal segmentation of [1.5, 6] seconds after each MI cue was extracted as one trial of EEG data.

\subsubsection{Dataset 2: MI (BCI4-2B)}
In BCI4-2B dataset\footnote{\url{www.bbci.de/competition/iv/\#dataset2b}}, EEG was collected with a 3 EEG electrode channels sampled at 250 Hz from nine healthy participants (ID B01-B09) in five separate acquisition sessions. Each participant participated in two motor imagery tasks, including imagining the movement of left hand and right hand. The first two sessions contained 120 trials per session without feedback and the last three sessions contained 160 trials per session with a smiley face on the screen as feedback. As described in \cite{zhao2020deep}, all data in the first three sessions were used for training and the last two sessions were used for test. Temporal segmentation of [3, 7] seconds after MI cue was extracted as one trial of EEG data in our experiment.

\subsubsection{Dataset 3: MI (BCI3-4a)}
In BCI3-4a dataset\footnote{\url{www.bbci.de/competition/iii}}, EEG was collected with a 118 EEG electrode channels sampled at 1000Hz from 5 healthy participants (ID: aa, al, av, aw, ay) contained 280 trials. Each participant participated in 3 motor imagery tasks, including imaging the movement of left hand, right hand and right foot. Only cues for the classes ‘right hand’ and ‘foot’ are publicly available, so the number of classes in this dataset is 2. The original data was obtained at a sampling rate of 1000 Hz. In order to verify the decoding ability of the model under different sampling rate, the original data was down-sampled to 128 Hz. BCI3-4a is a small training dataset. Taking the participant 'ay' as an example, it has only 28 training samples, which is a great challenge for the training of neural network models. To this end, we redivided the training set and test set in BCI3-4a and verified the cross-participant decoding ability of the models. Data from each participant was selected as test set in turn, and the others were used for training. Temporal segmentation for 3.5s after MI cue was segmented as one trial of EEG data according to the experimental setup.

\subsubsection{Dataset 4: P300-Speller (Kaggle-ERN)}
In kaggle P300-Speller (Feedback error-related negativity (ERN)) dataset\footnote{\url{www.kaggle.com/c/inria-bci-challenge}}, EEG was collected with a 56 EEG electrode channels sampled at 600Hz from 26 healthy participants (ID S01-S26). A matrix of 36 possible items was presented to each participant to spell words using letters and numbers. By flashing screen items in group and in random order, each word item was selected one at a time. The goal of this dataset is to determine when the selected item is not the correct one by analyzing EEG signals after the participant received feedback. According to the requirements of the competition, all models were trained and tested under the condition of cross participant and imbalanced classes. It should be noted that the data provided by the competition had been down-sampled at 200Hz, so we didn't do additional down-sampling. The data division of training set and test set also followed the requirements of the Kaggle competition. Moreover, the data of P300-Speller is generally unbalanced in categories. In this dataset, the ratio of positive and negative samples in the training set is 159:385. Temporal segmentation for 1.25s after cue was extracted as one trial of EEG data in our experiment.

\subsection{LMDA-Net architecture}

The LMDA-Net consists of a benchmark network and two attention modules: the channel attention module and the depth attention module. The channel attention module aims to improve the information screening ability in the spatial dimension of EEG signals and the depth attention module further refines the information of high-dimensional EEG features in depth dimension. 
The channel attention module and the depth attention module can be integrated with any convolutional neural networks.

The framework of LMDA-Net is shown in Figure \ref{fig:lmda}.  The design philosophy of LMDA-Net is closely related to the essential character of EEG. Firstly, although traditional deep learning is developing towards more and more complex models [\cite{krizhevsky2017imagenet,simonyan2014very, szegedy2016rethinking, he2016deep}], based on the knowledge disclosed by the neuroscience [\cite{jensen2010shaping, pfurtscheller1999event, polich2007updating}], the characteristics of EEG signals are relatively simple. Therefore, the shallow network has sufficient feature extraction ability for EEG decoding. Secondly, the amount of EEG data is scare while neural network models are data hungry, so a lightweight network is more effective in reducing overfitting. Finally, the spatial resolution of EEG is low, and the information in the temporal and spatial dimension contained in EEG is not equivalent. Combining the above characteristics, we designed the LMDA-Net in Figure \ref{fig:lmda}. The following is a detailed introduction to each part of the LMDA-Net.

\subsubsection{Benchmark Network} \label{benchmark}

\begin{figure*}
\centerline{\includegraphics[width=\textwidth]{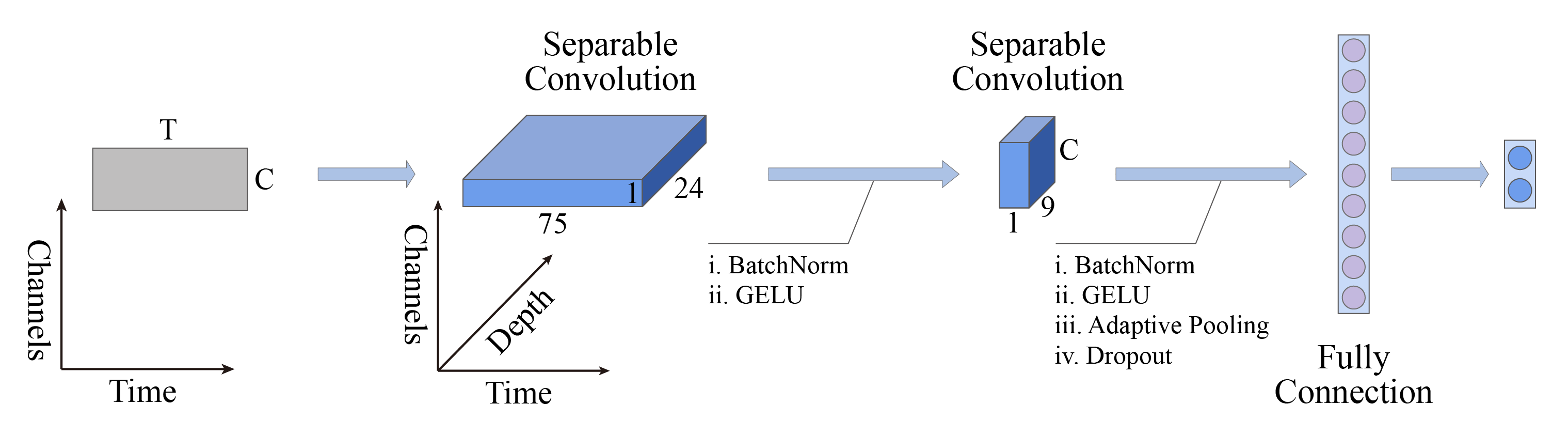}}
\caption{The architecture of the benchmark network.}
\label{fig:benchmark}
\end{figure*}

\begin{figure*}
\centerline{\includegraphics[width=\textwidth]{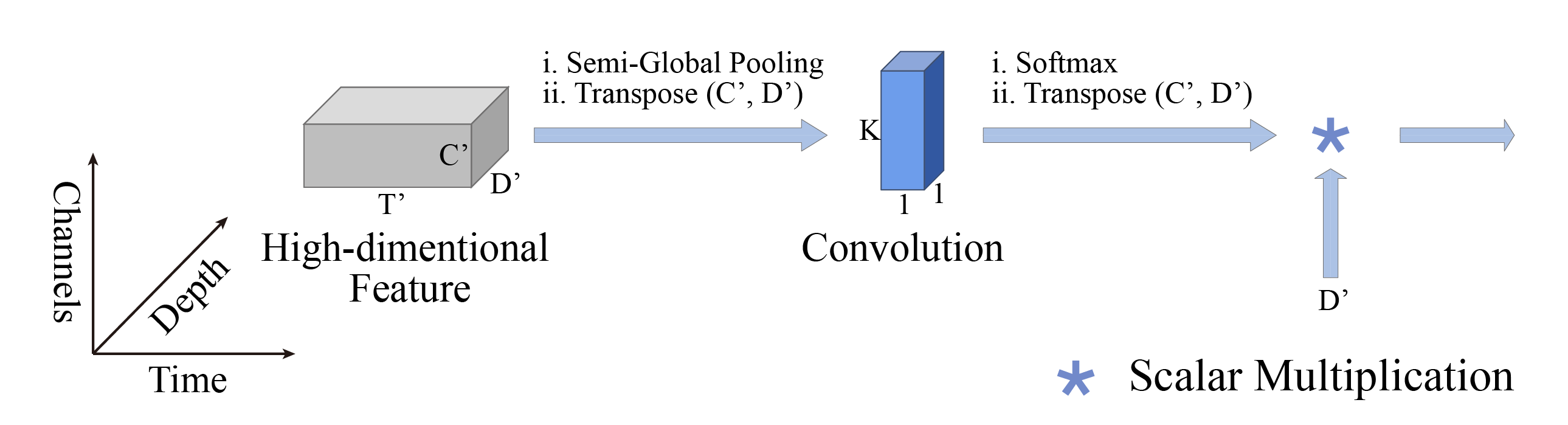}}
\caption{The architecture of the depth attention module, where T’ is the number of time samples, C’ is the number of channels, D’ is the number of kernels. Semi-global pooling is a two-dimensional pooling, which performs global average pooling on only one of the dimensions. The purpose of depth attention module is to strengthen the connection of features between temporal convolutional layer and spatial convolutional layer.}
\label{fig:depthattention}
\end{figure*}

The benchmark network of LMDA-Net is a combination of the main feature extraction architecture of ConvNet [\cite{schirrmeister2017deep}] and the separable convolution in EEGNet [\cite{lawhern2018eegnet}]. The feature extraction architecture of benchmark network consists of a temporal convolutional layer, a spatial convolutional layer, and a series of regularization methods. Each convolutional layer adopts a separable convolution to reduce the amount of network parameters. The size of the temporal convolutional kernel is set to (24, 1, 75), where 24 is the number of kernels, 1 and 75 are the size of convolutional kernel in spatial dimension and time dimension, respectively. And the size of the spatial convolutional kernel is set to (9, C, 1), where C is the number of EEG channels. Although fine-tuning the size of convolutional kernels according to the specific training set is helpful to improve the classification accuracy, the fixed parameters are more practical for evaluating the feature extraction ability of the neural network for EEG signals. For this reason, we have fixed the parameters in LMDA-Net in all experiments.

In addition, the activation function in the benchmark network is GELU [\cite{hendrycks2016gaussian}] , which has better smoothness than ELU [\cite{clevert2015fast}] used in EEGNet [\cite{lawhern2018eegnet}] and ConvNet [\cite{schirrmeister2017deep}].
Since fully connected layers create a large number of parameters to be trained, adaptive average pooling is designed to adaptively reduce the number of parameters of LMDA-Net according to different EEG tasks and acquisition devices. Adaptive average pooling is a two-dimensional pooling kernel, the size of the kernel is (1, $k_{pooling}$), where $k_{pooling}$ is shown in equation \eqref{eq:kpool}:
\begin{equation}
k_{pooling} = \max (1,\lfloor f / 10 / N\rfloor) \label{eq:kpool}
\end{equation}
where the operator $\lfloor \rfloor$ represents rounded down to a integer, $f$ is the sampling frequency of the input data, and $N$ is a value related to the number of training samples. According to EEG data collection experience, $ N=\max (1, \lfloor N_t /200\rfloor)$, where $N_t$ is the number of training samples.


Another difference from other networks [\cite{krizhevsky2017imagenet, lawhern2018eegnet, schirrmeister2017deep, simonyan2014very}] is that the proposed LMDA-Net decrease the number of convolution kernels from 24 to 9, as the network goes deeper. There are two main reasons for this. On the one hand, more kernels in the spatial convolutional layer will multiply the number of parameters in the fully connected layer, which makes the network quickly overfit and affect the feature extraction ability. On the other hand, the information contained in the time domain and the spatial domain of EEG signals is not equivalent. The time domain contains more information, so more convolution kernels are needed to extract features.

\subsubsection{Channel attention module} \label{channel_attention}

From the perspective of EEG acquisition, the electrical signal recorded by a single EEG electrode channel represents a superposition of multiple neuronal activities due to volume conduction effects. To improve the spatial resolution of EEG signals, some studies have utilized source reconstruction techniques to identify the neuronal regions that are associated with specific EEG activity for further analysis  [\cite{mammone2020deep},\cite{hou2020novel}]. However, this method requires strong prior knowledge and is not suitable for integration into end-to-end neural network models, making it challenging to decode EEG signals from multiple paradigms. From the perspective of neural network design, neural network models designed for EEG decoding, such as EEGNet [\cite{lawhern2018eegnet}], ConvNet [\cite{schirrmeister2017deep}] and DRDA [\cite{zhao2020deep}], perform temporal convolution first and then spatial convolution, which makes the temporal convolution layer lack of attention to the spatial information of EEG. 
Therefore, We propose a novel channel attention module to enhance the ability of neural networks to learn EEG spatial information, drawing inspiration from source reconstruction. 
The channel attention module acts on the input data and expands the spatial information of the input data to depth dimension by Tensor product.

Let $\mathbf{x}$ ($\mathbf{x} \in \mathbb{R}^{1 \times C \times T}$) be the input sample of EEG and $\mathbf{c}$ ($\mathbf{c} \in \mathbb{R}^{D \times 1 \times C }$) be a tensor following a normal distribution,  where $C$ is the number of channels, $T$ is the number of time samples,  and $D$ is the number of instances, numerically equivalent to the number of convolutional kernels.
With the help of Tensor product, the channel attention module maps the channel information of the input signal to the depth dimension, which integrates channel information with the subsequent temporal convolution. The equation of the channel attention module is given as follows:
\begin{equation}
    \mathbf{x}_{h c t}^{\prime}=\sum_{d} \mathbf{x}_{d c t} \mathbf{c}_{h d c} \label{eq2}
\end{equation}
The subscripts in equation \eqref{eq2} indicate the corresponding dimensions. The same letter of subscripts indicates that two tensors have the same shape in this dimension.

The number of parameters to be trained introduced in the channel attention module is D*C. A specific setting of the hyperparameter D for a specific task may perform better. However, to fairly compare the performance of LMDA-Net and other networks, the hyperparameter D is fixed to 9.
By introducing far fewer parameters than those of the benchmark network, the channel attention module  maps spatial information into the depth dimension, which provides a new idea for the attention mechanism of EEG signals.

\subsubsection{Depth attention module} \label{depth_attention}

In the field of computer vision, a feature map in the depth \footnote{In computer vision-related researches, the depth dimension is often expressed in terms of the channel dimension, which is the color channel of RGB. However, the word 'channel' is considered as electrode channel in EEG-related researches. To avoid confusion, We use the word 'depth' to represent the depth dimension.} dimension is considered as a feature detector [\cite{woo2018cbam}, \cite{zeiler2014visualizing}]. However, the features extracted from the temporal and spatial dimension of EEG are different. Feature maps in the depth dimension not only focus on ‘what’ is meaningful [\cite{woo2018cbam}] given an input, but also play the role of combining time and spatial features. Therefore, the depth attention methods in computer vision [\cite{woo2018cbam, hu2018squeeze, ecanet2020}] cannot be directly applied to EEG decoding. Taking CBAM [\cite{woo2018cbam}] as an example, the depth attention module in CBAM first aggregates depth information through global pooling, and then uses fully connected layers to further map and screen depth information. However, the feature map in EEG is not meaningful after global pooing and fully connected layers create a huge number of parameters to be trained, which will not only cause overfitting, but also affect the training effect of the benchmark network. For this reason, we designed a novel depth attention module suitable for EEG decoding inspired by the idea of ‘local cross-depth interaction' in [\cite{ecanet2020}]. The details of the depth attention module are shown in Figure \ref{fig:depthattention}. 

The depth attention module acts in the middle of the temporal convolution layer and the spatial convolution layer.
As shown in Figure \ref{fig:depthattention}, it is mainly composed of 3 parts: semi-global pooling, local cross-depth interaction and adaptive weighting. 
The semi-global pooling, which performs a global average pooling in the spatial dimension of the input feature $\mathbf{F}$ while retaining all the information in the time dimension, is first used to facilitate the screening of the depth information. And then, the depth information of the data is further screened by a convolutional layer. Compared with the fully connected layers, the convolutional layer can facilitate local interaction while reducing the number of parameters. 
Finally, the screened features are adaptively weighted. The depth information is first probabilized using the softmax function. Considering the sensitivity of EEG signals to amplitude changes, the output of the softmax function is further expanded to the same order of magnitude as the input features through Hadamard product.

Let the input feature of the depth attention module be $\mathbf{F} \in \mathbb{R}^{D^{\prime} \times C^{\prime} \times T^{\prime}}$. The depth attention module generates a map $\mathbf{M}(\mathbf{F}) \in \mathbb{R}^{D^{\prime} \times 1^{\prime} \times T^{\prime}}$ which can be represented by the following equation:
\begin{equation}
    \mathbf{M}(\mathbf{F})=\left(\textit{Softmax}\left(\textit{Conv}\left(\textit{Pooling}(\mathbf{F})^{T}\right)\right)^{*} D^{\prime}\right)^{T} \label{eq3}
\end{equation}
where \textit{Pooling} indicates the Semi-Global pooling, \textit{Conv} indicates the convolution operation and \textit{Softmax} indicates the softmax function. The superscript \textit{T} represents the transpose operation of the spatial dimension and the depth dimension of the tensor.

The number of parameters introduced in the depth attention module is k, where k is the number of the convolutional kernel parameters. Same as the channel attention module, a specific setting of the hyperparameter k for a specific task may perform better. The hyperparameter k is fixed to 7 to make a fair comparison with other networks in this paper.

\subsection{Experiment Settings}

\subsubsection{Preprocessing}
EEG signals were preprocessed with bandpass filtering and normalization, before feature extraction and classification. A 200-order Blackman windows bandpass filter was used to filter the raw EEG data.  Then the filtered raw EEG data was segmented according to the task duration in each dataset.
For MI task, the raw EEG data was bandpassed to [4, 38] Hz. And normalization methods, including channel normalization \cite{miao2022priming} and Euclidean alignment \cite{he2019transfer} were consistent with methods in \cite{miao2022priming}. For P300-Speller task, the cutoff frequencies of the bandpass filter are 1Hz and 40Hz \cite{lawhern2018eegnet}. And the baseline, which was the mean value of the data within 1s before cue, was subtracted from each trial. 

\subsubsection{Test criterions}
Neural network models often need a validation set to monitor the training process and select the optimal model for test. The validation set is usually obtained by sampling from the training set.
However, EEG has poor signal-to-noise ratio and a small number of data for training the neural network model. Such characteristics of EEG pose great challenge for validation set to determine the optimal model. For instance,  each participant in the BCI4-2A dataset has 288 training samples. If 20\% of the data is random sampled from the training set, the validation set will contain about 57 samples. Taking participant A01 as an example, we found in the experiment that the optimal model, monitored by the validation set composed of different samples, had a range of accuracy greater than 10\% in the test set. 
To avoid the bias of test results caused by different sampling methods of validation set, we trained all neural network models for 300 epochs and recorded the highest classification accuracy for the test set and the mean classification accuracy throughout the last 10 epochs. The highest classification accuracy represents the decoding ability of the model, and the average classification accuracy represents the predicting volatility of the model. Meanwhile, to simulate an online deployment scenario, the same initialization method was used in LMDA-Net for all participants.
For MI-related datasets, we also used kappa value \cite{zhao2020deep} as the evaluation metrics. For ERN dataset, due to class imbalance, AUC (Area Under the ROC Curve) was used instead of classification accuracy to evaluate the model.

\subsubsection{Experimental environment and parameter settings}
All the experiments were conducted under the Pytorch framework on a workstation with Intel(R) Xeon(R) Gold 5117 CPUs @ 2.00 GHz and Nvidia Tesla V100 GPUs. We trained all the models with AdamW as the optimizer, with the default parameters as described in \cite{loshchilov2018fixing} and mini batches with size of 32. 
It should be noted that ConvNet \cite{schirrmeister2017deep} has a shallow version and a deep version for EEG decoding, both EEGNet \cite{lawhern2018eegnet} and ConvNet \cite{schirrmeister2017deep} mentioned that shallow-ConvNet has more advantages than deep-ConvNet, thus we only selected shallow-ConvNet (Abbreviated as ConvNet below) in the experiment.

\section{Results}

\subsection{Experiment results}
\begin{figure*}
\centerline{\includegraphics[width=\textwidth]
{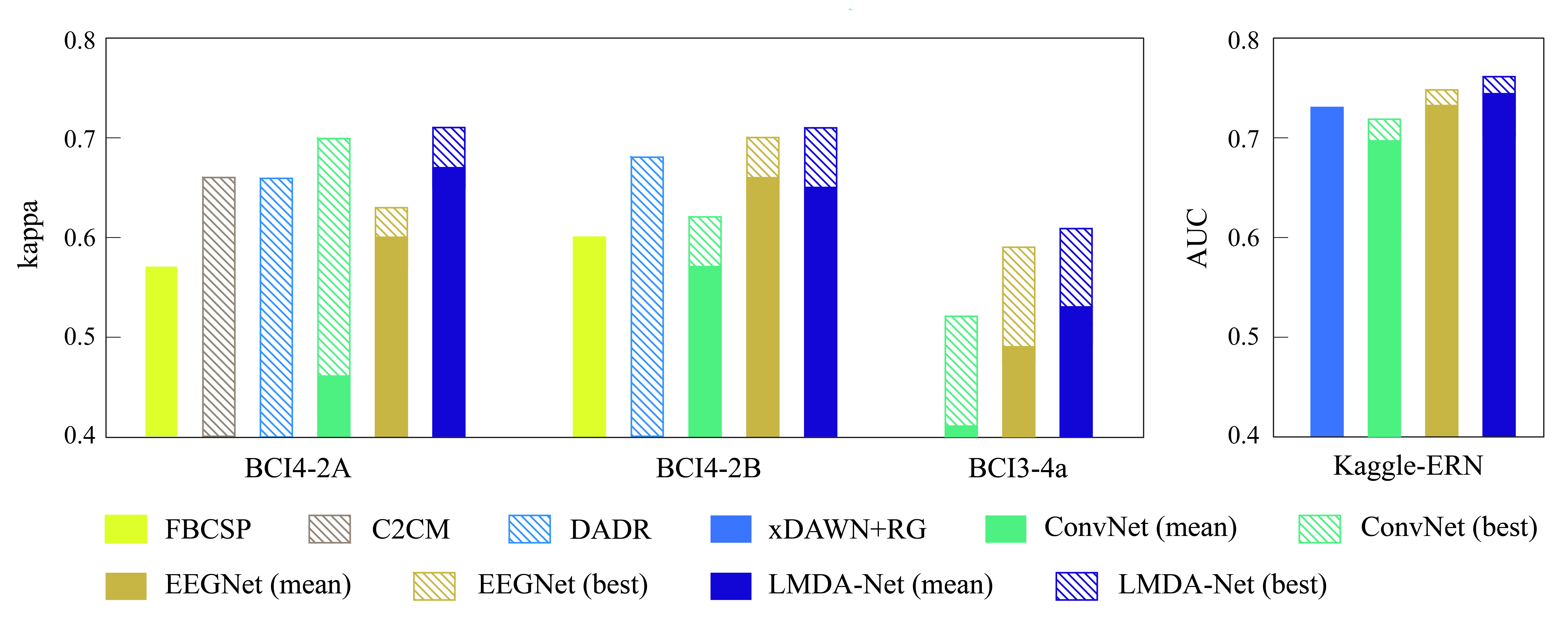}}
\caption{Classification performance of the competitive models on BCI4-2A, BCI4-2B, BCI3-4a and Kaggle-ERN, average across all participants. The results of manual feature extraction model are represented by an opaque solid column. The highest kappa value or AUC of neural network models is represented by a shaded column and the mean kappa value or AUC throughout the last 10 epochs is also represented by an opaque solid in the same column.    For BCI3-4a, the Kappa value of FBCSP is 0.2, not shown for overall layout. The kappa values of C2CM and DRDA are obtained from \cite{zhao2020deep}, which represent the highest accuary of the model, thus shown by the shaded columns as well.(Best viewed in color)}
\label{fig:results}
\end{figure*}

We tested the performance of LMDA-Net, EEGNet, ConvNet on the above datasets under the same conditions and also compared them to other representative methods.  
Figure \ref{fig:results} shows the classification performance for BCI4-2A, BCI4-2B, BCI3-4a and Kaggle-ERN datasets for the competitive models, average over all participants. More details about the classification results can be founded in the Appendix.

\subsubsection{BCI4-2A}
The hyperparameters in LMDA-Net were tuned via the BCI4-2A dataset. Therefore, we first evaluated the performance of different algorithms in the BCI4-2A dataset. As can be seen from Figure \ref{fig:results}, LMDA-Net shows highest classification accuracy (78.8\% accuracy and 0.71 kappa) and extremely low predicting volatility with mean 75.2\% accuracy (0.67 kappa) throughout the last 10 epochs, which represent that LMDA-Net is superior to other models in terms of classification performance. Under the same conditions, ConvNet shows higher classification accuracy (77.2\% accuracy, 0.7 kappa) than the EEGNet (72.6\% accuracy, 0.63 kappa). However, the higher decoding accuracy of ConvNet is at the expense of high predicting volatility.  The mean accuracy throughout epochs for ConvNet is much lower than that of EEGNet. This shows that ConvNet and EEGNet have their own advantages and disadvantages in model performance, which is also consistent with the conclusion in \cite{miao2022priming}.
Although DRDA uses data from other participants to assist training, with multi-dimensional attention mechanism to screen out EEG features, LMDA-Net improves the accuracy by 4.1\% compared with DRDA by using a small amount of training data and less training epochs. And C2CM sets the network architecture and parameters individually for each participant, LMDA-Net with fixed hyperparameters is still 4.4\% higher in the classification accuracy. 
FBCSP is the representative manual feature extraction algorithm of MI, and it was also the champion algorithm of this year's competition for this dataset. Compared with neural network models above, although FBCSP has the lowest computational cost, it no longer has the advantage in classification performance in BCI4-2A. This shows that the features mined by FBCSP are limited by the prior knowledge, and it also provides more possibilities for the neural network models in MI-based BCIs.

\subsubsection{BCI4-2B}
In the 2-class, 3-channel dataset, BCI4-2B, LMDA-Net also shows the highest classification accuracy (85.5\% acc, 0.71 kappa). Unlike in BCI4-2A, EEGNet in BCI4-2B shows higher classification performance (85.1\% acc, 0.70 kappa) than ConvNet (81.3\% acc, 0.62 kappa). In terms of predicting volatility, both LMDA-Net and EEGNet have  good performance and the mean accuracy throughout the epochs exceeds 82\% (0.64 kappa).  DRDA, which has a feature extraction network architecture similar to ConvNet, outperforms ConvNet in classification performance by drawing on data from other participants, but not as well as EEGNet and LMDA-Net. This shows that in the case of fewer EEG channels, the network with small parameters has more advantages in the classification performance. Compared with neural network models, FBCSP has no advantage in BCI4-2B. The possible reason is that it is difficult for the spatial filters  to mine effective features when the number of channels is small. It is worth noting that we do not optimize the  hyperparameters of LMDA-Net for BCI4-2B. The classification performance in BCI4-2B  reflects good robustness of LMDA-Net.

\subsubsection{BCI3-4a}
To further verify the robustness of LMDA-Net, we  conducted experiments on the BCI3-4a dataset. As shown in Figure \ref{fig:results}, LMDA-Net also shows the highest cross-participant classification accuracy (80.7\% acc, 0.61 kappa) and the low cross-participant predicting volatility. The mean accuracy throughout the last epochs of LMDA-Net achieves 76.5\% acc (0.53 kappa). This indicates that LMDA-Net also has advantages in MI classification under the condition of a large number of channels. In BCI3-4a, EEGNet has better performance than ConvNet. The possible reason is that there are too many EEG channels in this dataset. ConvNet, with a large number of parameters, is difficult to classify EEG signals with high complexity. The traditional method of FBCSP has less than 60\% accuracy in BCI3-4a, which is far lower than the neural network based methods. This shows that neural network methods have stronger robustness than FBCSP.

\subsubsection{Kaggle-ERN}
Compared with oscillatory motor imagery signals, signals of ERN are time-locked and thus have more obvious temporal features. So we continued to verify the performance of LMDA-Net on the Kaggle-ERN dataset. 
xDAWN+RG \cite{rivet2009xdawn, barachant2011multiclass, barachant2014plug} was the champion algorithm of this year's competition for this dataset. As can be seen from Figure \ref{fig:results}, its cross-participant classification performance also outperforms ConvNet. EEGNet, which has two layers of temporal convolution, also better cross-participant classification performance on Kaggle-ERN dataset.
Although LMDA-Net has only one layer of time-domain convolution, by introducing the channel-attention module and depth-attention module, LMDA-Net can also extract effective temporal features and achieve better classification performance (0.76 AUC) than EEGNet and xDAWN+RG. Meanwhile, LMDA-Net also has low predicting volatility, and the mean accuracy throughout the last 10 epochs of LMDA-Net is 0.74 AUC, which outperms xDAWN+RG with 0.73 AUC.

%
It can be seen from the above results that LMDA-Net outperforms  other competitive models on above four datasets in the classification performance, which shows that LMDA-Net has better generalization ability in EEG decoding for various tasks. Meanwhile, low predicting volatility and small training epochs of LMDA-Net makes it practical for real online BCI applications. 

\subsection{Ablation Study}

\begin{figure*}[!htb]
\centerline{\includegraphics[width=\textwidth]{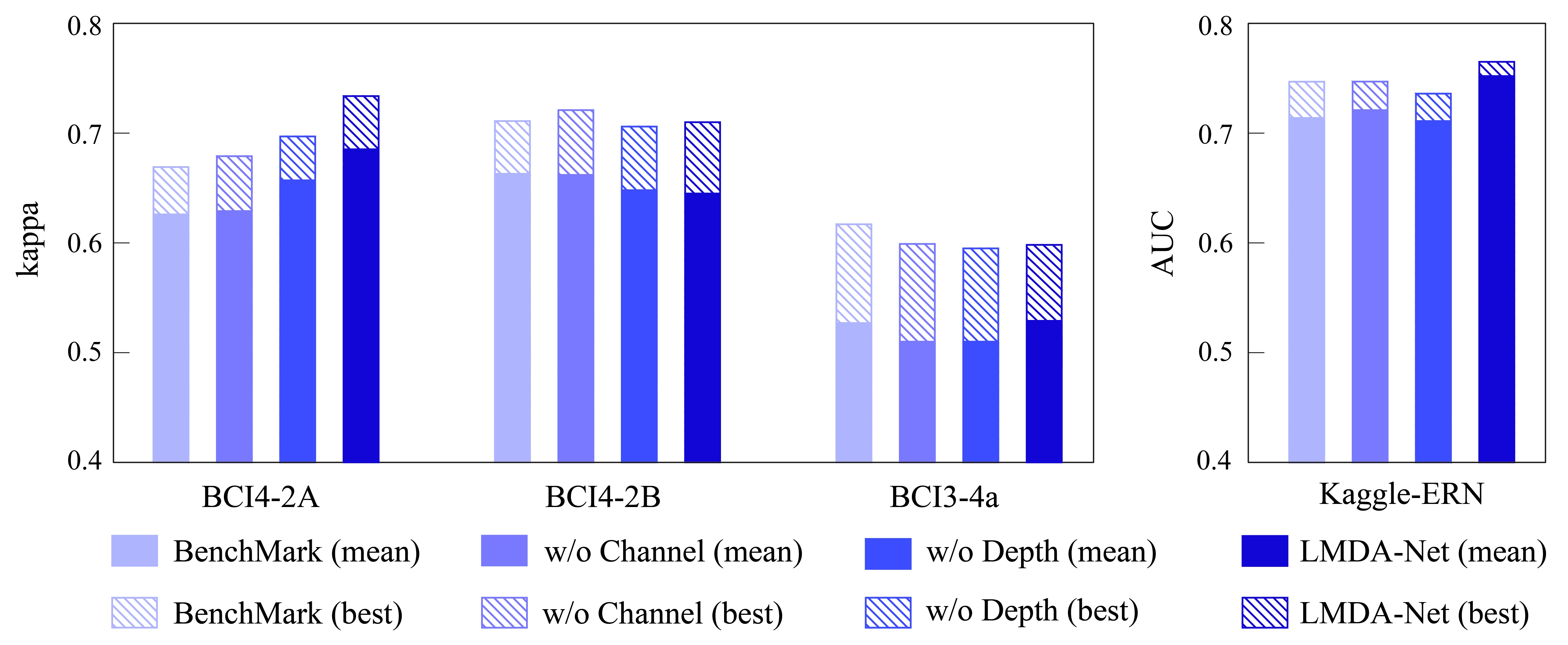}}
\caption{Ablation study on the structure of LDMA-Net. The bar chart shows the classification performance average across all BCI4-2A, BCI4-2B, BCI3-4a and Kaggle-ERN datasets. w/o Channel represents the LMDA-Net without channel attention module; w/o Depth represents LMDA-Net without depth attention module. The classification performance of the benchmark network is also presented for comparison. The highest accuracy of neural network models is shown by a shaded column, while the mean accuracy throughout the last 10 epochs is shown by an opaque solid column.(Best viewed in color)}
\label{fig:ablation}
\end{figure*}

We further performed ablation analyses on the above datasets to further investigate the effectiveness of channel attention module and depth attention module in LMDA-Net. We removed channel attention module and depth attention module in turn and compared them with LMDA-Net and the benchmark network.

\subsubsection{Benchmark network}
As the core architecture for feature extraction and classification of LMDA-Net, it is necessary to test the performance of benchmark network in different datasets separately. Combining the results shown in Figure \ref{fig:results} and Figure \ref{fig:ablation}, benchmark network integrates the advantages of EEGNet and ConvNet. In all MI datasets, the performance of the benchmark network outperforms EEGNet in both the highest classification accuracy and the mean classification accuracy throughout the last 10 epochs. Since there is only one layer of temporal convolution, the classification performance of benchmark network in Kaggle-ERN is less than that of EEGNet. The results in Kaggle-ERN indicate that features for ERN are more sensitive in the time domain, which is consistent with the phase-locked characteristics of P300. But compared to the ConvNet, which also has only one layer of temporal convolution, the performance of benchmark in Kaggle-ERN is greatly improved than that of ConvNet. In 3-channel BCI4-2B and 118-channel BCI3-4a dataset, the performance of benchmark network is almost the same as or even better than that of LMDA-Net, as shown in Figure \ref{fig:ablation}. This suggests that it is difficult to mine effective features when the spatial complexity of EEG is extremely low or high. Even so, due to the appropriate weights of the attention mechanism, it will hardly bring any loss to the network performance. However, in BCI4-2A and Kaggle-ERN datasets, the attention module on the benchmark can greatly improve the classification accuracy of the benchmark.

\subsubsection{Channel attention module}
Channel attention module is a module which is specially designed for EEG signals. To verify its impact on LMDA-Net, we removed the channel attention module of LMDA-Net and then tested the performance of the network. As shown in Figure \ref{fig:ablation}, in addition to BCI4-2B dataset, after removing the channel attention module, the performance of LMDA-Net drops significantly, which shows that the channel attention module is able to effectively regularize the model to extract effective features. In the 3-channel BCI4-2B dataset, after removing the channel attention module, the average accuracy across all participants reaches 0.72 kappa, outperforming all other models. Due to the low spatial information of BCI4-2B, using the channel attention module to screen just three EEG channels would be ineffective. On the contrary, using the depth attention module to screen depth information can gain additional benefits for BCI4-2B. 

\subsubsection{Depth attention module}
According to Figure \ref{fig:ablation}, after removing the depth attention module, the performance of LMDA-Net in all datasets has decreased to a certain extent, indicating that the depth attention module is an inseparable part of the LMDA-Net. The impact of the depth attention module on MI and P300-Speller is slightly different. Classification performance of LMDA-Net on the Kaggle-ERN dataset is more sensitive to the depth attention module. When removing the depth attention module from LMDA-Net, the performance of the network is similarly to that of the benchmark network. This suggests that the deep attention module can compensate for the deficiency of single-layer temporal convolution in LMDA-Net for feature extraction of P300-Speller signal. Compared with adding an additional layer of temporal convolution, the depth attention module uses fewer parameters and is more applicable for multiple paradigms.

As shown in Figure \ref{fig:ablation}, the classification results of BCI4-2B and BCI3-4a do not change much in different models during the ablation experiments, and the classification accuracy depends heavily on the performance of the benchmark network. The possible reason is that the spatial information complexity of EEG data in these two datasets is either extremely low or high. When the number of channels is too small, it is meaningless to screen the spatial information. At this time, it is recommended to use the depth attention module individually to assist the training of benchmark network. When the spatial complexity of EEG data is too high, complex models are no longer advantageous in classification without data dimension reduction. However, the integration of attention modules in LMDA-Net introduces an effective weighting method. In the BCI4-2A and Kaggle-ERN datasets, the integration of the channel attention module and the depth attention module has greatly improved the classification performance of the benchmark network. This also makes the performance of LMDA-Net outperforms EEGNet and ConvNet in all the tested four datasets, making LMDA-Net a more suitable network model for EEG based BCI applications.

\subsection{Interpretability}

\begin{figure*}[htbp]
\centerline{\includegraphics[width=\textwidth]{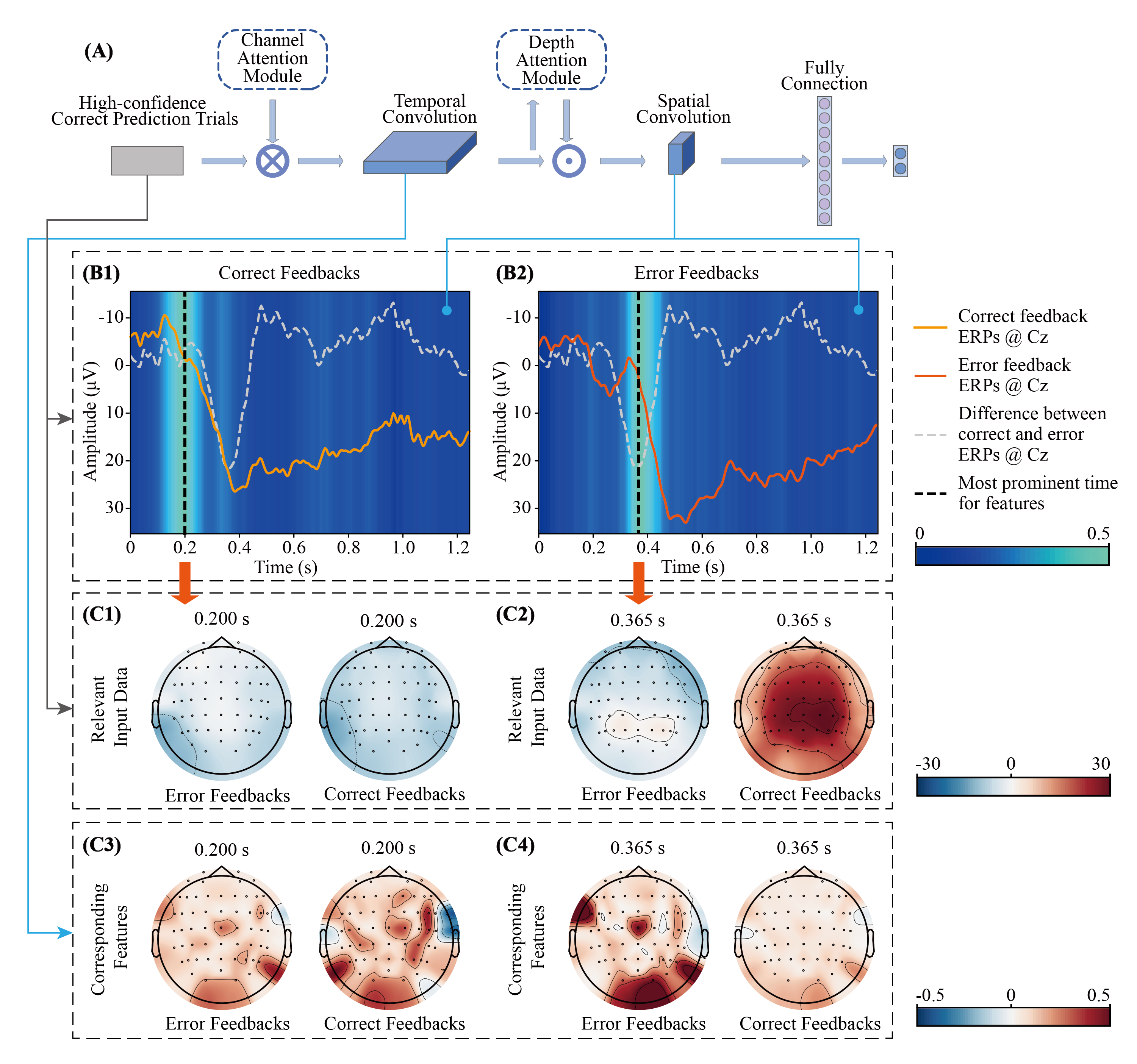}}
\caption{Comparison with the temporal and spatial features with ERPs and LMDA-Net visualized by Eigen-CAM. (A) The sketch of the well-trained cross-participant LMDA-Net for Kaggle-ERN dataset. High-confidence correct prediction trials in each test participant (approximately 9 trials per class per participant) are selected as the input data. The temporal visualization is computed by the features on the spatial convolutional layer, using Eigen-CAM, while the spatial visualization is computed by the features on the temporal convolutional layer. The grey line and light blue line from the sketch of LMDA-Net indicate the data source for the plot, which points out whether the plot is drawn from the input data or features extracted by LMDA-Net. (B1) and (B2) show the average ERPs at Cz channel (shown as the curve graph) and visualized features from the spatial convolutional layer (shown as the background) of correct and error feedback trials, respectively, in the temporal domain, averaged across all the test participants. Zero time corresponds to the feedback onset. The maximum value computed by Eigen-CAM in the temporal domain is marked with a black dashed line and denoted as the most prominent time. (C1) and (C2) show the brain topographies of the input data from correct and error feedback trials, respectively, averaging across all participants, at the most prominent time. (C3) and (C4) show the visualization of features from temporal convolutional layer for correct and error feedbacks at the most prominent time, respectively. For ease of comparison, (C3) and (C4) also map the visualization results in the spatial domain to the brain topographies in the same way as (C1) and (C2).(Viewed in color)}
\label{fig:ern}
\end{figure*}

\begin{figure*}[htbp]
\centerline{\includegraphics[width=\textwidth]{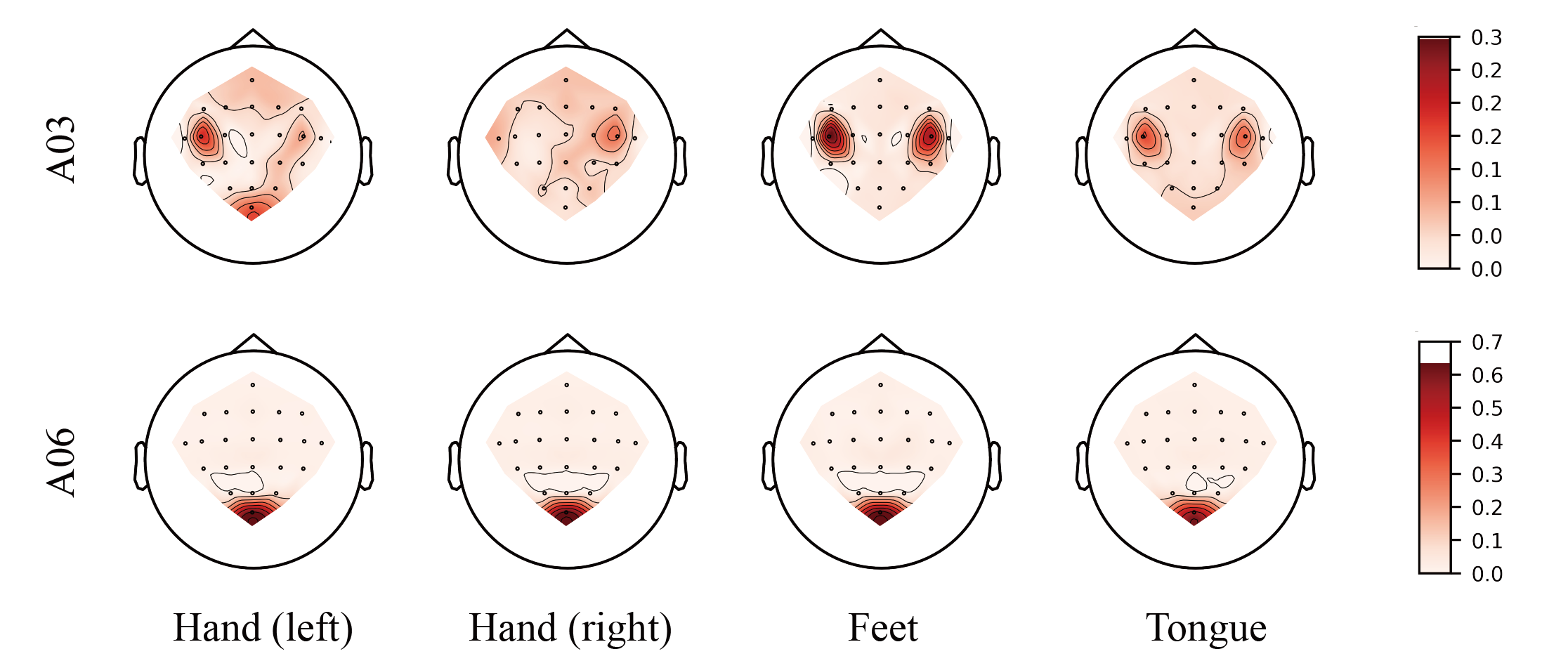}}
\caption{Spatial feature visualization of within-participant trained LMDA-Net models for participant A03 and A06 of the BCI4-2A dataset. This is a frequency-independent class activation map for channels in the spatial dimension. The specific implementation method is shown in Algorithm \ref{alg2}. (Best viewed in color)}
\label{fig:mi}
\end{figure*}

Class activation map (CAM) is a technique to visualize the regions of input data that are important for predictions from CNN-based models \cite{chattopadhay2018grad}, \cite{selvaraju2017grad}. CAM is mainly used in the field of computer vision, which utilizes features learned from the convolutional layer to produce a coarse localization map of the important regions in the input image. Among CAM tools \cite{chattopadhay2018grad,li2016visual,muhammad2020eigen,selvaraju2017grad}, Eigen-CAM \cite{muhammad2020eigen} computes and visualizes the principal components of the learned features and can localize objects robustly and reliably without the need to modify model architecture or backpropagation of gradients. It can be seen from \cite{muhammad2020eigen} that Eigen-CAM show better capability of localizing the most discriminative regions compared to other methods. So we chose Eigen-CAM as the visualization tool to locate the features learned by LMDA-Net. 

The ERN is a negative deflection in the ERPs and is attributed primarily to activities in the anterior cingulate cortex (ACC). This kind of event has distinct temporal and spatial characteristics. To this end, We proposed Algorithm \ref{alg1} for interpretability analysis of LMDA-Net's features for ERN-related signals. 

To clarify the temporal and spatial locations that LMDA-Net payed attention to when extracting features from the Kaggle-ERN dataset, all the training data in Kaggle-ERN dataset are used to train a cross-participant model. Then the trained model is used to predict all the trials of each test participant in turn and select high-confidence trials with correct predictions (approximately 9 trials per class per participant) as the input data for visualization.  Finally, the results in Figure \ref{fig:ern} can be obtained by Algorithm \ref{alg1}.

\begin{algorithm}[htbp]
	\renewcommand{\algorithmicrequire}{\textbf{Input:}}
	\renewcommand{\algorithmicensure}{\textbf{Output:}}
	\caption{Feature interpretability of LMDA-Net for ERN-related signals}
	\label{alg1}
	\begin{algorithmic}[1]
        \REQUIRE High-confidence correct prediction trials $\mathbf{D}_{ij} \in \mathbb{R}^{N\times{C}\times{T}}$; corresponding label $y$ 
        \ENSURE  (1) Feature visualization of LMDA-Net in the time domain\\
                 (2) Feature visualization of LMDA-Net in the spatial domain\\
                 (3) ERPs at Cz channel\\
                 (4) brain topographies of ERN \\
        \STATE \# \textit{$i$:index of test participant; $j$: index of class; $N$:number of high-confidence correct prediction trials for each participant per class; $C$: number of channels; $T$: number of time samples;} \\
        \# \textit{Feature visualization for LMDA-Net}
        \FOR{$i \leftarrow 1 \text{ to } 10 $ \text{ and } $j \leftarrow 0 \text{ to } 1 $}     \STATE get the class activation map value  $\mathbf{S}_{ij} \in \mathbb{R}^{N\times{C}\times{T}}$ and $\mathbf{T}_{ij} \in \mathbb{R}^{N\times{C}\times{T}}$ by Eigen-CAM, corresponding to $\mathbf{D}^{ij}$, generated from LMDA-Net's spatial convolutional layer and temporal convolutional layer, respectively \\
            \STATE \# \textit{average over trials}
            \STATE $\mathbf{S}_{ij}^{\prime} = \text{mean}(\mathbf{S}_{ij}, \text{dim}=0)$,
            $\mathbf{S}_{ij}^{\prime} \in \mathbb{R}^{\text{1}\times{C}\times{T}}$ \\
            $\mathbf{T}_{ij}^{\prime} = \text{mean}(\mathbf{T}_{ij}, \text{dim}=0)$,
            $\mathbf{T}_{ij}^{\prime}\in\mathbb{R}^{\text{1}\times{C}\times{T}}$
            \STATE \# \textit{average over EEG channels}
            \STATE $\mathbf{T}_{ij}^{\prime\prime} = \text{mean}(\mathbf{T}_{ij}^{\prime}, \text{dim}=1)$,
            $\mathbf{T}_{ij}^{\prime\prime} \in \mathbb{R}^{\text{1}\times{\text{1}}\times{T}}$
        \ENDFOR
        \FOR{$j \leftarrow 0 \text{ to } 1 $} 
        \STATE \# \textit{get the class activation map value in the time domain corresponding to each class}
        \STATE $\mathbf{t}_{j}^{\prime\prime} = \sum_{i}\mathbf{T}_{ij}^{\prime\prime}$, $\mathbf{t}_{j}^{\prime\prime} \in \mathbb{R}^{T}$ \\
        \STATE \# \textit{get the background image of Figure 6 (B1) and (B2)} \\
        \STATE map $\mathbf{t}_{j}^{\prime\prime}$ into a heat map
        \STATE \# \textit{average over all participants}
        \STATE $\mathbf{S}_{j}^{\prime}=\frac{1}{N}\sum_{i}\mathbf{S}_{ij}^{\prime}$, $\mathbf{S}_{j}^{\prime} \in \mathbb{R}^{\text{1}\times{C}\times{T}}$
        \STATE get the time index $t_j$ (the most prominent time for features) of the maximum value of $\mathbf{T}_{j}^{\prime\prime}$
        \STATE get the value of $\mathbf{S}_{j}^{\prime}$ at time $t_j$,  denoted by $\mathbf{s}_{j}^{\prime\prime}$, $\mathbf{s}_{j}^{\prime\prime} \in \mathbb{R}^{C}$ \\
        \STATE \# \textit{get Figure 6 (C3) and (C4)} 
        \STATE map $\mathbf{s}_{j}^{\prime\prime}$ into the brain topography \\
        \# \textit{Feature Visualization in Neuroscience}
        \STATE \# \textit{average over trials}
        \STATE $\mathbf{D}_{ij}^{\prime} = \text{mean}(\mathbf{D}_{ij}, \text{dim}=0)$,
            $\mathbf{D}_{ij}^{\prime} \in \mathbb{R}^{\text{1}\times{C}\times{T}}$
        \STATE \# \textit{average over participants}
        \STATE $\mathbf{D}_{j}^{\prime}=\frac{1}{N}\sum_{i}\mathbf{D}_{ij}^{\prime}$, $\mathbf{D}_{j}^{\prime} \in \mathbb{R}^{\text{1}\times{C}\times{T}}$
        \STATE select the value of $\mathbf{D}_{j}^{\prime}$ corresponding to channel Cz, $\mathbf{d}_{j}^{\prime}$, $\mathbf{d}_{j}^{\prime} \in \mathbb{R}^{T} $
        \STATE \# \textit{get the amplitude-time graph of Figure \ref{fig:ern} (B1) and (B2)}
        \STATE plot the amplitude-time graph of $\mathbf{d}_{j}^{\prime}$ 
        \STATE get the value of $\mathbf{D}_{j}^{\prime}$ at time $t_j$, denoted by $\mathbf{d}_{j}^{\prime\prime}$, $\mathbf{d}_{j}^{\prime\prime} \in \mathbb{R}^{C}$ \\
        \STATE \# \textit{get Figure \ref{fig:ern} (C1) and (C2)}
        \STATE plot the brain topography according to $\mathbf{d}_{j}^{\prime\prime}$
        \ENDFOR
	\end{algorithmic}  
\end{algorithm}

\begin{algorithm}
	\renewcommand{\algorithmicrequire}{\textbf{Input:}}
	\renewcommand{\algorithmicensure}{\textbf{Output:}}
	\caption{Feature interpretability of LMDA-Net for BCI4-2A dataset}
	\label{alg2}
	\begin{algorithmic}[1]
        \REQUIRE High-confidence correct prediction trials $\mathbf{D}_{j} \in \mathbb{R}^{N\times{C}\times{T}}$ for the participant; corresponding label $y$ 
        \ENSURE  Feature visualization of LMDA-Net in the spatial domain
        \STATE \# \textit{The subscript has the same meaning as in Algorithm 1} 
        \FOR{$j \leftarrow 0 \text{ to } 3 $}     
        \STATE get the class activation map value  $\mathbf{S}_{j} \in \mathbb{R}^{N\times{C}\times{T}}$ and $\mathbf{T}_{j} \in \mathbb{R}^{N\times{C}\times{T}}$ by Eigen-CAM, corresponding to $\mathbf{D}^{ij}$, generated from LMDA-Net's spatial convolutional layer and temporal convolutional layer, respectively \\
        \STATE  \# \textit{average over EEG channels}
        \STATE $\mathbf{T}_{j}^{\prime} = \text{mean}(\mathbf{T}_{j}, \text{dim}=1)$, $\mathbf{T}_{j}^{\prime} \in \mathbb{R}^{N\times{\text{1}}\times{T}}$
        \STATE select the time index corresponding to the first 10 maximum values for each trial in $\mathbf{T}_{j}^{\prime}$, denoted by $S_t = \{s_1, s_2, ...s_N\}$
        \STATE get the value of $\mathbf{S}_{j}$ under the index of $S_t$, denoted by $\mathbf{S}_{j}^{\prime}$, $\mathbf{S}_{j}^{\prime} \in \mathbb{R}^{N\times{C}\times{\text{10}}}$
        \STATE  \# \textit{squeeze the dimension of time samples}
        \STATE $\mathbf{S}_{j}^{\prime\prime}=\sum(\mathbf{S}_{j}^{\prime}, dim=2)$, $\mathbf{S}_{j}^{\prime\prime} \in \mathbb{R}^{N\times{C}}$
        \STATE  \# \textit{get the weight of each channel}
        \STATE convert $\mathbf{S}_{j}^{\prime\prime}$ into a probability distribution on dimension 1.
        \STATE  \# \textit{eliminate the effect of the number of trials}
        \STATE $\mathbf{s}_{j}^{\prime\prime}=mean(\mathbf{S}_{j}^{\prime\prime}, dim=0)$, $\mathbf{s}_{j}^{\prime\prime} \in \mathbb{R}^{C}$
        \STATE  \# \textit{get Figure 7}
        \STATE map $\mathbf{s}_{j}^{\prime\prime}$ into the brain topography
        \ENDFOR
	\end{algorithmic}  
\end{algorithm}


Figure \ref{fig:ern} (B1) and (B2) show the time course of ERPs at channel Cz and the result of temporal feature visualization of LMDA-Net for the selected correct and error feedback trials, respectively. For ease of comparison, the results of temporal feature visualization have the same time coordinates as the ERPs curve, and are stretched longitudinally and displayed as the background of ERPs curves. 

The maximum value of temporal feature visualization is marked with a black dashed line in Figure \ref{fig:ern} (B1) and (B2). It can be seen that the results of temporal feature visualization are related to the white dashed line, which reflects the difference between the ERPs corresponding to correct feedbacks and error feedbacks at channel Cz. In \cite{margaux2012objective}, the positive component around 0.365s recorded by channel Cz is called pos-ErrP, and the first negative component ahead of the positive component is called neg-ErrP. Although there are differences between the dataset used in this paper and \cite{margaux2012objective}, the similar time course of ERPs in different datasets also indicate that time-domain features are the key to distinguish different classes of ERN signals.
As shown in Figure \ref{fig:ern} (B1) and (B2), the focuses of feature visualization in the time domain are also on these two components. The time corresponding to the peak of pos-ErrP is 0.365s, which is exactly the time corresponding to the maximum value of temporal feature visualization of error feedbacks (target stimuli). 
From the results of temporal feature visualization, LMDA-Net's focus in the time domain coincides with the components of N1, N2, and P300, which indicates that the discriminative features learned by LMDA-Net are consistent with the prior knowledge in neuroscience.
Figure \ref{fig:ern} (C1) and (C2) are the topographies of the input signal at the most prominent time, which are used as comparison for (C3) and (C4). Figure \ref{fig:ern} (C1) shows that topographies of the input signal have little difference for different events at 0.2s, while the topographies have obvious differences at 0.365s, which can also be indirectly seen from the white dotted line in Figure \ref{fig:ern} (B2). However, the topographies of the input data cannot reflect the importance of channels. In neuroscience, a common method is to select channels with significant differences between target and non-target stimuli within a specific time window. For example, in \cite{margaux2012objective}, channel Cz, P7, and P8 are selected to study the time response of ERN respectively. As the results of temporal feature visualization are channel-independent, the importance of channels under different events can be revealed at the prominent time. Figure \ref{fig:ern} (C3) and (C4) show the spatial feature visualization of LMDA-Net at the prominent time. For ease of comparison, the spatial feature visualization is also carried out using the same method of topography in MNE \cite{GramfortEtAl2013a}. Although there is little difference in topographies at 0.2s between the different classes of input signals, as shown in Figure \ref{fig:ern} (C1), the spatial feature visualization in (C3) shows LMDA-Net has obvious attention to the correct feedback event in channel P7. Figure \ref{fig:ern} (C4) shows a significant differentiation of spatial feature visualization betweent error feedbacks and correct feedbacks, and channel Cz, F7, P8 and O2 are paid high attention from LMDA-Net when identifying error feedback events. These channels of interest are also highly consistent with those commonly used in neuroscience, which reflects the plausibility of LMDA-Net for spatial feature extraction.

The signal of motor imagery is known as the sensorimotor rhythm (SMR) which is not phase-locked to the event. The common method for spatial mapping of SMR is to display the significant ERD/ERS in time-frequency maps for each motor imagery task. And then select a distinctive frequency band in a time window to display the topographical distribution of ERD/ERS. \cite{pfurtscheller1999event, pfurtscheller2006mu}
To obtain spatial feature visualization of LMDA-Net, we further proposed Algorithm \ref{alg2}. 
Because of the large individual variability of motor imagery signals among participants, we selected two representative participants, A03 and A06, from BCI4-2A dataset for the analysis. The former has good classification performance in both manual feature extraction algorithms and automatic feature extraction algorithms, while the latter has consistently poor classification performance. Meanwhile, EEGNet \cite{lawhern2018eegnet} also performed a spatial visual analysis of A03, so the results of spatial feature visualization of different methods can be compared.

We selected top 5 high-confidence correct prediction trials in the test data for each task per participant as the input data in Algorithm \ref{alg2} and obtained Figure \ref{fig:mi}. As shown in Figure \ref{fig:mi}, the result of spatial feature visualization of A03 indicates that LMDA-Net mainly focuses on C3 and C4 channels when classifying.  Unlike spatial mapping of ERD/ERS, LMDA-Net extracts discriminative features among different tasks. Thus the results of spatial feature visualization reflect the weight of channels among different MI tasks. For different tasks, LMDA-Net pays attention to different channels, which indicates good separability in spatial domain for the data in A03.  And these channels often show significant ERD/ERS for different MI tasks in neuroscience.  Although Algorithm \ref{alg2} does not perform additional time-frequency analysis on the input signal and does not select a specific time window, the spatial visualization results of A03 are similar to \cite{pfurtscheller2006mu}(using different datasets), which further demonstrates the rationality of Algorithm \ref{alg2}. 
Compared with EEGNet \cite{lawhern2018eegnet}, which visualizes the spatial mapping of spatial filters, Algorithm \ref{alg2} can display class-specific visualization results. And instead of abstract spatial mapping of filters, the result of Algorithm \ref{alg2} are highly correlated with classification, thus providing better interpretability for SMR rhythm. 

Algorithm \ref{alg2} can also provide spatial feature interpretability for BCI illiteracy, like A06. As shown in Figure \ref{fig:mi}, LMDA-Net's spatial attention among different tasks is all concentrated in channel POz and has a higher degree of attention compared with other channels. This indicates poor separability in spatial domain for the data in A06. Moreover, POz's spatial visualization value in A06 also far exceeds the value of the most significant channels in A03. We further examined the raw EEG data and found that the values of POz are significantly larger than those of other channels due to some undesired factors, which makes Poz become the dominant channel and has negative effects in classification. 


\section{Conclusion}
In this paper, we propose a general lightweight multi-dimensional attention network, named LMDA-Net to handle EEG decoding across multiple paradigms. Channel attention module and depth attention module, are proposed and validated for EEG signals to effectively extract features and jointly integrate high-dimensional features across different dimensions. By introducing a small number of parameters, two attention modules improve the decoding ability of benchmark network in both MI and P300-speller. Experimental results on 4 public datasets show that LMDA-Net achieves the highest classification accuracy and low predicting volatility in all tested datasets compared with other representative methods, which suggests LMDA-Net is an all-purpose EEG deep learning architecture appliable to multiple paradigms. In addition, to solve the interpretability problem of the neural network in EEG recognition, we further propose neural network visualization algorithms suitable for both evoked responses and endogenous activities. The visualization results are not only consistent with the prior knowledge in neuroscience, but also provide an intuitive explanation for the classification accuracy. In summary, LMDA-Net is an efficient feature extraction network and has the potential to be an important corner stone for EEG-based BCI applications.



\section*{Appendix}

\begin{table*}[hbtp]
\caption{classification accuracy (\%) of different algorithms on BCI4-2A}
\centering
\label{tabel:2a}
\begin{tabular}{|c|c|c|c|c|c|c|c|c|c|c|}
\hline
         & A01           & A02           & A03           & A04           & A05           & A06           & A07           & A08           & A09           & Average acc$\pm$std (kappa)  \\ \hline
\cite{ang2008filter}    & 76.0          & 56.5          & 81.2          & 61.0          & 55.0          & 45.2          & 82.7          & 81.2          & 70.7          & 67.7$\pm$13.7(0.57$\pm$0.18)          \\ \hline
\cite{sscsp}     & 76.7          & 58.6          & 81.2          & 57.6          & 38.5          & 48.2          & 76.3          & 79.1          & 78.8          & 66.1$\pm$15.8(0.54$\pm$0.21)          \\ \hline
\cite{ssmm}    & 82.6          & 60.7          & 85.7          & 67.0          & 58.6          & 54.5          & 90.9          & 81.2          & 78.5          & 73.4$\pm$13.3(0.64$\pm$0.18)          \\ \hline
\cite{lawhern2018eegnet}   & 75.3          & 51.0          & 88.5          & 57.3          & 46.5          & 50.3          & 83.7          & 80.5          & \textbf{87.1}          & 68.9$\pm$17.4(0.58$\pm$0.23)          \\ \hline
\cite{schirrmeister2017deep}  & 76.4          & 55.2          & 89.2          & 74.6          & 56.9          & 54.1          & \textbf{92.7}          & 77.1          & 76.4          & 72.5$\pm$14.2(0.63$\pm$0.19)          \\ \hline
\cite{sakhavi2018learning}   & \textbf{87.5} & 65.2 & 90.2 & 66.6          & 62.5 & 45.4          & 59.5          & \textbf{88.3} & 79.5          & 74.4$\pm$15.5(0.65$\pm$0.21)          \\ \hline
\cite{zhao2020deep}    & 83.1          & 55.1          & 87.4          & 75.2          & 62.2          & 57.1          & 86.1          & 83.6          & 82.0          & 74.7$\pm$13.0(0.66$\pm$0.17)          \\ \hline
LDMA-Net (ours) & 86.5          & \textbf{67.4}          & \textbf{91.7}          & \textbf{77.4} & \textbf{65.6}          & \textbf{61.1} & 91.3 & 83.3          & 85.4 & \textbf{78.8}$\pm$11.5(\textbf{0.71}$\pm$0.15) \\ \hline
\end{tabular}
\end{table*}

\begin{table*}[hbtp]
\caption{classification accuracy (\%) of different algorithms on BCI4-2B}
\centering
\label{tabel:2b}
\begin{tabular}{|c|c|c|c|c|c|c|c|c|c|c|}
\hline
         & B01           & B02           & B03           & B04           & B05           & B06           & B07           & B08           & B09           & Average acc$\pm$std (kappa)  \\ \hline
\cite{ang2008filter}   & 70.0          & 60.3          & 60.9          & 97.5          & 93.1          & 80.6          & 78.1          & 92.5          & 86.8          & 80.0$\pm$13.9(0.60$\pm$0.28)          \\ \hline
\cite{ssmm}    & 74.0          & 55.0          & 55.6          & 94.0          & 86.8          & 82.1          & 76.5          & 92.1          & 85.6          & 78.0$\pm$14.4(0.56$\pm$0.29)          \\ \hline
\cite{sscsp}    & 65.0          & 56.7          & 54.0          & 95.6          & 74.6          & 79.0          & 80.0          & 87.8          & 82.8          & 75.0$\pm$14.0(0.50$\pm$0.28)          \\ \hline
\cite{lawhern2018eegnet}  & 77.5          & 61.0          & 63.1          & \textbf{98.4} & \textbf{96.5} & 83.7          & 84.3          & 92.8          & 88.4          & 82.9$\pm$13.5(0.65$\pm$0.27)          \\ \hline
\cite{schirrmeister2017deep} & 74.3          & 56.0          & 57.5          & 97.5          & 95.3          & 82.1          & 79.6          & 87.5          & 86.5          & 79.6$\pm$14.8(0.59$\pm$0.30)          \\ \hline
\cite{zhao2020deep}    & \textbf{91.3} & \textbf{62.8} & 63.6          & 95.9          & 93.5          & 88.1          & \textbf{85.0}          & \textbf{95.2}          & 90.0          & 83.9$\pm$12.8(0.67$\pm$0.26)          \\ \hline
LDMA-Net (ours) & 81.2          & 62.1          & \textbf{71.8} & \textbf{98.4} & 95.6 & \textbf{89.3} & \textbf{85.0} & 94.6 & \textbf{91.8} & \textbf{85.5}$\pm$12.0(\textbf{0.71}$\pm$0.24) \\ \hline
\end{tabular}
\end{table*}

\begin{table*}[hbtp]
\caption{classification accuracy (\%) of different algorithms on BCI3-4a}
\centering
\label{tabel:4a}
\begin{threeparttable}
\begin{tabular}{|c|c|c|c|c|c|c|}
\hline
         & aa            & al            & av            & aw            & ay            & Average acc$\pm$std (kappa)  \\ \hline
\cite{ang2008filter}   & 54.6          & 71.4          & 54.6          & 63.9          & 51.7          & 59.2$\pm$8.2(0.18$\pm$0.16)          \\ \hline
\cite{lawhern2018eegnet}  & \textbf{81.8} & 87.1          & \textbf{78.2} & 83.3          & 66.0          & 79.3$\pm$8.1(0.58$\pm$0.16)          \\ \hline
\cite{schirrmeister2017deep}  & 74.1          & 86.6          & 67.3          & 82.8          & \textbf{71.2} & 76.4$\pm$8.1(0.52$\pm$0.16)          \\ \hline
LDMA-Net (ours) & 80.3          & \textbf{93.9} & 74.6          & \textbf{88.5} & 66.4          & \textbf{80.7}$\pm$10.9(\textbf{0.61}$\pm$0.22) \\ \hline
\end{tabular}
    \begin{tablenotes}
        \footnotesize
        \item The same preprocessing methods and training strategies are used in EEGNet and ConvNet as in LMDA-Net.
    \end{tablenotes}
\end{threeparttable}
\end{table*}

Table \ref{tabel:2a}, \ref{tabel:2b}, \ref{tabel:4a} 
show the detailed results for each participant under different algorithms on BCI4-2A, BCI4-2B, BCI3-4a and Kaggle-ERN datasets, respectively. We chose algorithms with the same test context and without data augmentation to fairly compare the accuracy of each participant under different algorithms. 
LMDA-Net used the same parameters and initialization methods in all four datasets.
It is more suitable for online decoding for diverse EEG tasks. Here are some differences between other models and LMDA-Net:
\begin{enumerate}
\item C2CM set different network parameters for each participant on BCI4-2A dataset, while the parameters of LMDA-Net were designed to accommodate all participants in the four datasets.
\item DRDA used the data from other participants to assist the target participant in training, while LMDA-Net used only the training data of the target participant.
\end{enumerate}

\section*{Statistics and reproducibility}  
We used the hold-out test set method in all experiments and fixed all initialization seeds to ensure the reproducibility of the results obtained from the neural network model. 
Meanwhile, we calculated the mean accuracy (or AUC) of the prediction results throughout the last 10 epochs (290-299 epochs) for each training to analyze the volatility of the model. This will  provide a rough measure of how well the model performs in an online system.
For motor  imagery datasets, we also calculated the average accuracy and variance for each participant due to the large individual differences. 

\section*{Data and code availability statement}
The raw EEG data that support the findings of this study are available in \url{www.bbci.de/competition/iv/\#dataset2a}, \url{www.bbci.de/competition/iv/\#dataset2b},  \url{www.bbci.de/competition/iii} and \url{www.kaggle.com/c/inria-bci-challenge}, respectively.  

The source code for EEGNet and xDAWN+RG is publically available at the following webpage: \url{https://github.com/vlawhern/arl-eegmodels}. The source code for ConvNet is publiclly available at \url{https://github.com/TNTLFreiburg/braindecode}. We are pleased to provide publiclly source code for LMDA-Net at \url{https://github.com/MiaoZhengQing/LMDA-Code}.

\section*{Credit authorship contribution statement}
\textbf{Zhengqing Miao}: Conceptualization, Methodology, Formal analysis, Investigation, Visualization, Writing -original draft. \textbf{Xin Zhang} Validation, Writing –review\& editing, Funding acquisition and Supervision. \textbf{Meirong Zhao} Supervision, Funding acquisition, Writing –review\& editing. \textbf{Dong Ming} Supervision, Project administration, Writing –review\& editing.

\section*{Competing Interests statement}
The authors declare no competing interests. 

\section*{Acknowledgments}
This work was supported in part by the National Key Research and Development Program of China under Grant 2022YFF1202900, the National Natural Science Foundation of China under Grant 82102174, China Postdoctoral Science Foundation under Grant 2021TQ0243, and Tianjin Science and Technology Planning Project under Grant 20JCQNJC01250. 
We would like to thank Yang Li for her work in collecting the datasets and Yiwei Yang for touching up the figures.

\clearpage
\bibliographystyle{cas-model2-names}
\bibliography{LMDA_references}

\end{document}